\documentclass[journal,twoside,web]{ieeecolor}
\usepackage{tmi}
\usepackage{cite}
\usepackage{amsmath,amssymb,amsfonts,bbm}
\usepackage{bbding}
\usepackage{algorithmic}
\usepackage{graphicx}
\usepackage{textcomp}
\usepackage{multirow}
\usepackage{overpic}
\usepackage{subfigure}
\usepackage{threeparttable}
\usepackage{cases}

\def\BibTeX{{\rm B\kern-.05em{\sc i\kern-.025em b}\kern-.08em
    T\kern-.1667em\lower.7ex\hbox{E}\kern-.125emX}}
\begin{document}
\title{Medical Matting: Medical Images Segmentation with Uncertainty from Matting Perspective}
\author{
	Lin Wang, \IEEEmembership{Student Member, IEEE},
	Lie Ju,
	Xin Wang, 
	Wanji He,
	Donghao Zhang,	
	Yelin Huang,
	Zhiwen Yang,
	Xuan Yao,
	Xin Zhao,
	Xiufen Ye, \IEEEmembership{Senior Member, IEEE},
	and Zongyuan Ge \IEEEmembership{Senior Member, IEEE}
\thanks{(Corresponding author: Zongyuan Ge, Xiufen Ye)}
\thanks{Lin Wang and Xiufen Ye are with Harbin Engineering University, Harbin, Heilongjiang 150001, China. Lin Wang is also with the Monash-Airdoc joint research group, Monash University, Clayton, VIC 3800 Australia (e-mail: wanglin.mailbox@gmail.com, yexiufen@hrbeu.edu.cn).}
\thanks{Lie Ju, Donghao Zhang, and Zongyuan Ge are with Monash University, Clayton, VIC 3800 Australia. Lie Ju and Zongyuan Ge are also with Airdoc, Beijing 100089, China (e-mail: julie334600@gmail.com, {donghao.zhang, zongyuan.ge}@monash.edu).}
\thanks{Xin Wang, Wanji He, Yelin Huang, Zhiwen Yang, Xuan Yao, and Xin Zhao are with Airdoc, Beijing 100089, China (e-mail: {wangxin, hewanji, huangyelin, yangzhiwen, yaoxuan, zhaoxin}@airdoc.com).}
\thanks{Lie Ju, Xin Wang, Wanji He, and Donghao Zhang contributed equally.}}

\maketitle

\begin{abstract}

It is difficult to accurately label ambiguous and complex shaped targets manually by binary masks. The weakness of binary mask under-expression is highlighted in medical image segmentation, where blurring is prevalent. In the case of multiple annotations, reaching a consensus for clinicians by binary masks is more challenging. 
Moreover, these uncertain areas are related to the lesions' structure and may contain anatomical information beneficial to diagnosis. However, current studies on uncertainty mainly focus on the uncertainty in model training and data labels. None of them investigate the influence of the ambiguous nature of the lesion itself. 
Inspired by image matting, this paper introduces alpha matte as a soft mask to represent uncertain areas in medical scenes and accordingly puts forward a new uncertainty quantification method to fill the gap of uncertainty research for lesion structure. 
In this work, we introduce a new architecture to generate binary masks and alpha mattes in a multitasking framework, which outperforms all state-of-the-art matting algorithms compared.
The proposed uncertainty map is able to highlight the ambiguous regions and a novel multitasking loss weighting strategy we presented can improve performance further and demonstrate their concrete benefits.
To fully-evaluate the effectiveness of our proposed method, we first labelled three medical datasets with alpha matte to address the shortage of available matting datasets in medical scenes and prove the alpha matte to be a more efficient labeling method than a binary mask from both qualitative and quantitative aspects.

\end{abstract}


\label{sec:introduction}

\begin{figure}[t]
	\begin{center}
		\begin{overpic}[width=\columnwidth]{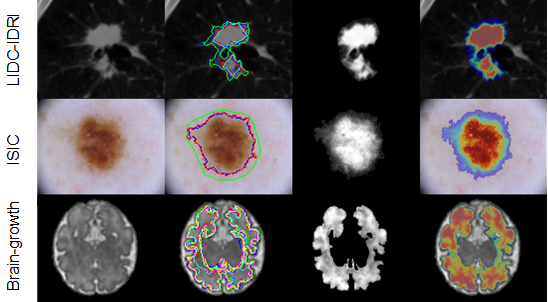}
			\put(17,-3.5){(a)}
			\put(40,-3.5){(b)}	
			\put(63,-3.5){(c)}
			\put(86,-3.5){(d)}
		\end{overpic}
	\end{center}
	\caption{Examples of the ambiguity in medical images. (a) Original images of lung nodule~\cite{armato2004lung}, skin lesions~\cite{codella2018skin}, and newborn brain myelination process~\cite{https://qubiq.grand-challenge.org}. (b) Contours of the binary manual labels (distinguished by colors). Inaccuracy and inconsistency are more likely to be located in the ambiguity regions. (c) Targets labeled by the alpha matte. It is more structural descriptive than binary masks by showing sufficient information about the edges and the internal texture. (d) Original images mixed with the alpha matte in pseudo-color for better display.}
	\label{fig:ambiguity}
\end{figure}
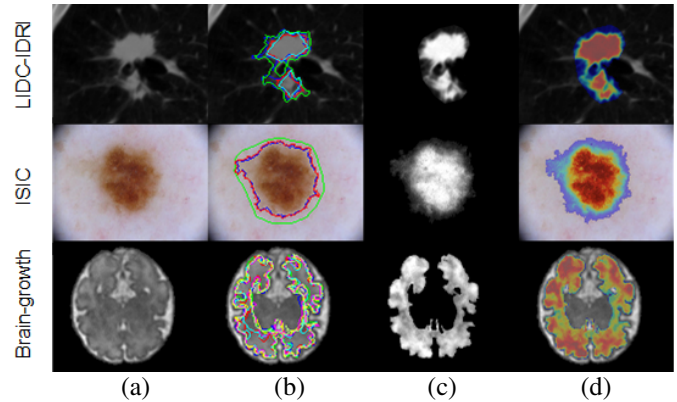

\IEEEPARstart{A}{mbiguities} are common in medical image segmentation, and eliminating ambiguities is a challenge. 
From a practical application point of view, it includes image blurring due to imaging principles, low resolution limited by the imaging systems, progressive manifestations of lesion development, etc., shown in Fig.~\ref{fig:ambiguity}-(a).
Besides, the labeling process can also introduce semantic ambiguity. 
For example, variation of clinicians' experience may result in disagreements in annotations, and sometimes limited by annotation tools, small lesion structures can only be labelled with rough outlines. Furthermore, the binary mask labels cannot accurately describe the transition region, resulting in information loss. 

However, these fuzzy and hard-to-segment areas are of great diagnostic values, which is better represented by continuous labels rather than binary labels. 
In terms of the lung nodule dataset~\cite{armato2004lung}, the uncertain regions can provide unique value and better describe the indistinct border and ground-glass shadow around a lesion, which is vital for nodule staging. Similarly, in the skin lesion dataset~\cite{codella2018skin}, the lesions' fuzzy boundary and internal structures are used to triage and diagnose dermatoses. In the Brain-growth dataset~\cite{https://qubiq.grand-challenge.org}, the newborn's white tissue undergoes a rapid myelination process and thus it is difficult to label the white matter in a shifting region myelinated or non-myelinated with a binary value. So the continuous labelled region is useful to track the disease progression. 

Due to the diagnostic values of uncertain regions, quantifying these regions with annotation ambiguity is in great need. Fig.~\ref{fig:ambiguity}-(b) demonstrates the difficulty in accurately describing lesions with binary mask due to the image's fuzziness and the ambiguous transition region from the targeting region to the healthy region, which in most of the time leads to annotation disagreements.
From the data side, some existing works attempt to reduce the uncertainty by mapping multiple annotations into binary value~\cite{armato2004lung,codella2018skin,https://qubiq.grand-challenge.org}, but this do not address the limitation of using certain binary masks to describe uncertain lesions and tends to resulting in mislabelling. 

There are also some efforts being made to quantify uncertainties under the deep learning frameworks~\cite{lakshminarayanan2017simple,kendall2017bayesian,kendall2017uncertainties,rupprecht2017learning,kohl2018probabilistic,baumgartner2019phiseg,hullermeier2019aleatoric}. However, these methods aim at the uncertainties in models or labels and usually quantify them by multi-sampling, such as ensemble methods, suffering from high training resource consumption. Moreover, as far as we know, there is no research dedicated to exploring the inherent uncertainty of lesion structure in medical images.

In order to overcome the shortcomings of the binary mask, this work proposes a new kinds of image-matting based segmentation method~\cite{aksoy2017designing,chen2013knn,chuang2001bayesian,levin2007closed,wang2008image,xu2017deep,lutz2018alphagan,cai2019disentangled,forte2020fbamatting} using continuous values to distinguish and quantify the uncertain lesions, which pays attention to the uncertainties related to the characteristics of lesions and has better representation capability than simple binary masks.
Image matting is a specific segmentation approach widely used in picture editing, green screening, virtual conference~\cite{xu2017deep,cai2019disentangled}, etc. 
Matting assumes that the image $\mathcal{I}$ is a mixture of foreground $\mathcal{F}$ and background $\mathcal{B}$. The mixing coefficients, named alpha matte $\boldsymbol{\alpha}$, of foreground and background can be obtained through the matting operation, which makes: 
\begin{equation}
	\mathcal{I}=\boldsymbol{\alpha}\mathcal{F}+(1-\boldsymbol{\alpha})\mathcal{B},
	\label{eq:matting}
\end{equation}
where each element is in $\boldsymbol{\alpha}\!\in\![0, 1]$.
There are many approaches to Matting, including conventional closed-form solutions~\cite{aksoy2017designing,chen2013knn,chuang2001bayesian,levin2007closed}, deep learning methods~\cite{xu2017deep,lutz2018alphagan,cai2019disentangled,forte2020fbamatting}, and we refer readers to~\cite{wang2008image,xu2017deep} for a more comprehensive understanding of the detailed definition and challenges of image matting task.

\begin{figure}[t]
	\begin{center}
		\begin{overpic}[width=\columnwidth]{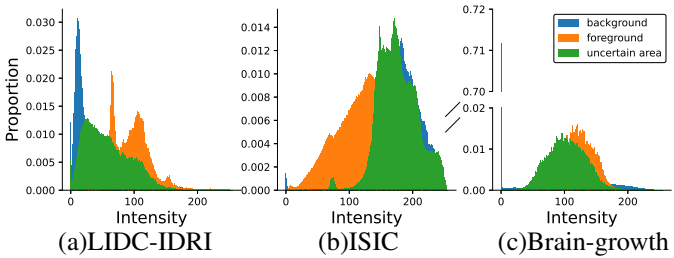}
			\put(8,-3.5){(a)LIDC-IDRI}
			\put(47,-3.5){(b)ISIC}	
			\put(74,-3.5){(c)Brain-growth}
		\end{overpic}
	\end{center}
	\caption{Pixel intensity distributions in foreground, background, and uncertain areas of the datasets. 
	The distributions of the uncertain area having a large overlap with both the foreground and background distributions, implying that the uncertain area contains both the foreground and background information.
	}
	\label{fig:distribution}
\end{figure}

Analogically, we can use the degree of mixing the pathological tissue and its surrounding normal tissue as a soft mask to describe the lesion in more detail.
So the diseased lesion corresponds to the foreground $\mathcal{F}$ in Eq.~\ref{eq:matting}, while the normal tissue would correspond to the background $\mathcal{B}$. 
The inconsistency area among the multiple annotations are taken as uncertain area, since it lies between the absolute foreground and the background zone, implying that the pixels are in a transition zone. We visualized the intensity distributions of foreground, background, and uncertain area in the three multi-annotated datasets, as shown in Fig.~\ref{fig:distribution}. 
The uncertainty related to the characteristics of lesions can be defined as entropy,
\begin{equation}
	\mathcal{U}=-\boldsymbol{\alpha}\log \boldsymbol{\alpha} - (1-\boldsymbol{\alpha})\log(1-\boldsymbol{\alpha}).
\end{equation} 
where the uncertainty is lower for pixels that are more certain to be lesions or normal tissue, and vice versa.
We illustrate the results of the alpha matte in Fig.~\ref{fig:ambiguity}-(c) and (d). It can be observed that using the continuous labelling values can better characterize and indicate the anatomical structure of lesions than binary masks. 

In this paper, we introduce matting to medical images with a multi-task learning network as a finer segmentation method while establishing a connection with the uncertainty reflecting the characteristics of the lesion itself.
Our main contributions can be summarized as follows:
\begin{itemize}
	\item Alpha matte is first being introduced in medical scenarios to address the challenge of uncertainty region representation and quantification. More valuable information in ambiguity regions are reserved. A novel content-related uncertainty quantification method is proposed accordingly.
	\item An integrated end-to-end multi-task network is proposed, producing the alpha matte and binary masks simultaneously. Uncertainty map is generated based on the predictions of binary masks, which improves the performance of the matting network and plays a similar role as trimap used in image matting.
	\item We labelled and released three datasets of various medical modalities to study uncertainty region segmentation problem. Datasets, labels and codes are publicly available as benchmark materials for the research community\footnote{Url for codes and datasets: \color{blue}{https://github.com/wangsssky/MedicalMatting}}.
\end{itemize}

This work extends our MICCAI 2021 paper~\cite{10.1007/978-3-030-87199-4_54}
in the following new contributions: 
\begin{itemize}
	\item A novel multitasking loss weighting strategy is proposed to achieve a coarse-to-fine optimization by dynamically and steadily switching the training focus, further improving the model's performance.
	\item Quantitative experiments have been carried out for more comprehensive ablation studies on the key components, including uncertainty map,  matting network, and multitasking loss weighting strategies, and elucidate their contribution to the network.
	Moreover,  down-streaming experiments on skin lesion classification illustrate the benefits of the alpha matte in diagnosis.	
	\item A new dataset constructed by using dermatoscopy images from the ISIC 2018 dataset~\cite{codella2018skin} is added, and the effectiveness of our method is verified on this dataset, which also demonstrates the robustness and generalisation capability of our proposed method to various medical imaging modalities (CT, MRI, dermoscopy).
\end{itemize}

The paper is structured as follows: Section~\ref{sec:related work} provides background information of uncertainty learning and matting in medical scenarios. Section~\ref{sec:datasets} introduces the datasets for this work. Section~\ref{sec:methods} illustrates components of the medical matting model. Section~\ref{sec:experiments} presents the experiments and ablation study.
In Section~\ref{sec:conclusion}, the potential use of medical matting in diagnosis, limitations and future work are discussed.


\section{Related works}\label{sec:related work}
\subsection{Uncertainty learning}
The accuracy of segmentation is important in medical diagnosis, but it is not easy to achieve perfection in practice. Thus, understanding the confidence of prediction, i.e., the uncertainty, becomes a practical solution.

Many deep-learning-based methods are proposed to quantify the uncertainty and alleviate its impact on diagnosis and prognosis, such as ensemble of models~\cite{lakshminarayanan2017simple}, Monte Carlo sampling with dropout~\cite{kendall2017bayesian}, multi-heads model~\cite{rupprecht2017learning}.
But basically, their idea is to simulate the results generated by multiple models and then evaluate the uncertainty by the variability between them, which can be described by 
    $\mathcal{U}=f(\mathcal{P})$, 
where $\mathcal{P}$ is a set of segmentation predictions and $f(\cdot)$ is a operation to calculate the variance or entropy.

Some methods try to produce results that fit a certain distribution through generative models.
Kohl~\cite{kohl2018probabilistic} proposed a generative segmentation model named Probabilistic UNet (Prob. UNet) by combining a UNet~\cite{ronneberger2015u} with a conditional variational autoencoder~\cite{kingma2014auto} and learning the distribution of the multi-annotated labels with the prior and posterior network structures, which can produce an unlimited number of binary masks.
Baumgartner~\cite{baumgartner2019phiseg} and Kohl~\cite{kohl2019hierarchical} proposed hierarchical probabilistic networks for modeling the segmentation at different resolutions, which can model multi-scale ambiguities and outperforms the Prob. UNet in generating more realistic and diverse segmentation samples.
Moreover, uncertainty methods can be used in the refinement process. Soberanis-Mukul~\cite{soberanis2020uncertainty} computed the uncertainty by applying Monte Carlo dropout, then deployed a graph convolution neural network to refine the uncertain region, which showed a better result than UNet and conditional random field based refinement.

However, most of these previous approaches define uncertainty by learning differences in prediction results by different means, which mainly target uncertainty in the training model and data labels without considering the uncertainty due to structural natures of lesions in images and possible labeling biases caused by them. 
In this paper, we combine the segmentation of medical images with matting, and use alpha matting to construct an uncertainty quantification method related to lesion structures. 

\subsection{Matting in medical images}

Image matting is not widely used in medical scenes 
and most of the related methods~\cite{zeng2012region,levin2007closed,cheng2017awm} focus on refining the mask boundary to improve segmentation performance. Another common usage of matting in medical images is to construct a trimap as an auxiliary for finer manipulation to uncertain regions~\cite{fan2018hierarchical,zhao2020improving,kim2021uacanet}.   

Matting can better extract and represent the edge information in the image to be applied to edge optimization of segmentation results. 
Zeng~\cite{zeng2012region} introduced the Closed-Form Matting~\cite{levin2007closed} into tumor segmentation of each slice of the PET images of head-and-neck cancer patients.
They regarded the tumor and normal region as foreground and background, respectively, and the pixels on the boundary as a mixture of foreground and background. 
The matting operation was used to refine the inexact boundary generated by segmentation.
Similarly, Cheng~\cite{cheng2017awm} derived adaptive weight matting from a local regression and global alignment view for gray-scale images and applied it to the medical image segmentation task. The evaluation results on a limited dataset showed that this method is comparable to the Graph Cut and Closed-Form Matting in the Dice Similarity. 
	
Trimap can provide prior information of image components, such as foreground, background, and uncertain areas, thus reducing the complexity of the matting task. This mechanism is also applied to some medical segmentation tasks to improve performance further. 
Considering the continuity and extendibility of retinal vessels, Fan~\cite{fan2018hierarchical} proposed a matting model for retinal vessel segmentation.
The unknown pixels in uncertain regions were divided into several hierarchies and sorted according to their distance to the nearest foreground pixel. 
Then, the unknown pixels were iteratively labelled as foreground or background by the spatial and color correlations with the nearest labelled pixel. 
Zhao~\cite{zhao2020improving} generated a trimap for the matting module by thresholding the score map of the segmentation network with two thresholds, such that the network can focus more on the region which could be wrongly identified. Then, a learned global threshold was applied to the matting output to obtain a binary mask. 
Kim~\cite{kim2021uacanet} computed a trimap and aggregated it with the feature map as a context guiding mechanism in each bottom-up stream prediction module of the UNet structure, which achieved state-of-the-art in polyp segmentation.  

Although the preceding matting applications in medical scenes demonstrate the promise of matting in medical imagery, practically all of them focus on improving segmentation accuracy rather than fully exploiting matting's expression capabilities in the semantics of the fuzzy nature. The proposed method integrates matting technology with uncertainty analysis, promoting the use of matting in the medical field.

\section{Datasets}\label{sec:datasets}

Three multi-annotated datasets with alpha mattes are used in this work, covering CT, MRI, and optical images. Specifically, they are a subset of LIDC-IDRI~\cite{armato2004lung} and the Brain-growth of QUBIQ~\cite{https://qubiq.grand-challenge.org}, and also a part of ISIC 2018 dataset~\cite{codella2018skin}. 
The first two were introduced in our MICCAI paper and reviewed by clinicians. The third one is newly added to this extension. 

The LIDC-IDRI dataset includes thoracic CT scans for lung nodules diagnosis. 
By convention~\cite{hu2019supervised,kohl2018probabilistic}, the scans are cropped and centered to a size $128\times128$.
Each patch is labelled out the region of the pulmonary nodules by four binary masks.
To better focus on the uncertainty study, we selected the patches in which an identical nodule is labelled in the corresponding masks. 
The Brain-growth consists of low-intensity contrast T2-W MR images for the white matter tissue myelination process of the newborn brain.
The ISIC dataset contains dermoscopic images intended for skin lesion segmentation and identification.
Table~\ref{tab:datasets} shows the total number images, the number of annotators, the image format, the image type, and the medical matting targets of each dataset.

\begin{table}[!t]
	\caption{Details of the datasets.}
	\begin{center}
		\addtolength{\tabcolsep}{-2pt}
		\begin{tabular}{l|ccc}
			\hline
			Datset	&LIDC-IDRI~\cite{armato2004lung}   &ISIC~\cite{codella2018skin}  &Brain-growth~\cite{https://qubiq.grand-challenge.org} \\			
			\hline
			\#Image	&1609	&120	&39 \\
			\#Annotators	&4	&3	&7 \\			
			Format	&Grayscale	&Color	&Grayscale \\
			Type	&CT	&Dermoscope &MRI \\
			Target	&Pulmonary nodule	&Melanoma / Nevi	&Myelination process \\
			\hline										 
		\end{tabular}
	\end{center}
	\label{tab:datasets}
\end{table}

\begin{figure}[t]
	\begin{center}
		\begin{overpic}[width=\columnwidth]{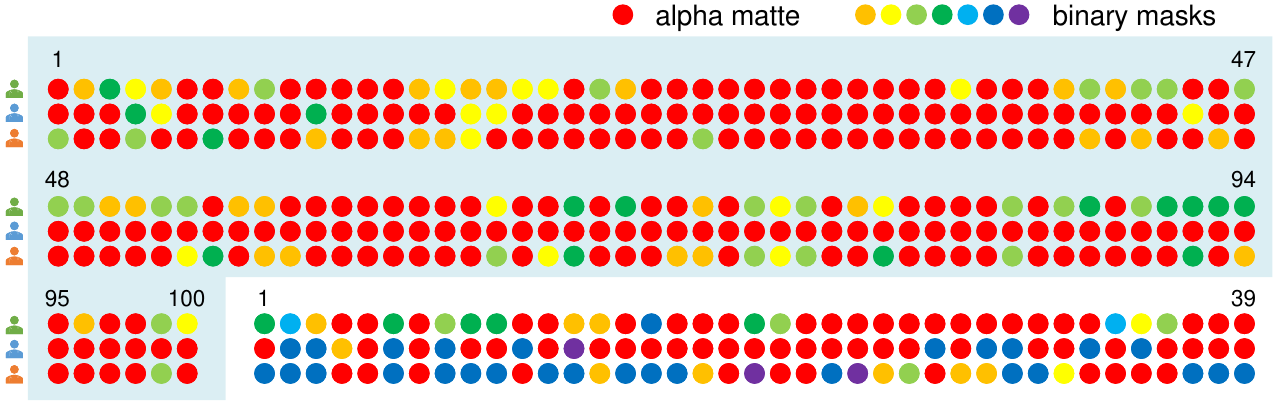}
			\put(6,-3.5){(a)}
			\put(55,-3.5){(b)}	
		\end{overpic}
	\end{center}
	\caption{The most descriptive label (the alpha matte or one of the binary masks) for each medical image evaluated by three clinicians. 
	We evaluated a hundred randomly selected cases of the LIDC-IDRI (a) and the whole cases of the Brain-growth (b), where rows are votes of the clinicians, and columns are different image cases. 
	From the proportion of red dots, the alpha mattes are more preferred in expressing the anatomical structures than the binary masks.}
	\label{fig:vote}
\end{figure}

\begin{table}[!t]
	\caption{Clinicians' preference for alpha matte versus binary masks evaluated by t-test.}
	\begin{center}
		\begin{tabular}{l|ccc|ccc}
			\hline
			P-val.&	 \multicolumn{3}{c|}{LIDC-IDRI} & \multicolumn{3}{c}{Brain-growth}\\
			\hline
			Mask	&\#C1   &\#C2  &\#C3 	&\#C1 	&\#C2 	&\#C3 \\			
			
			\#M1	&6.5E-08	&1.3E-95	&1.5E-22	&6.7E-08 	&1.7E-12 	&\textcolor{red}{0.056} \\
			\#M2	&1.8E-12	&7.2E-83	&2.6E-32 	&5.5E-10	&4.5E-14	&0.00064 \\ 
			\#M3	&6.5E-08	&1.3E-95	&1.0E-25	&6.7E-08	&4.5E-14 	&0.00064 \\ 
			\#M4	&8.9E-12	&2.1E-73	&4.0E-28 	&2.5E-06 	&4.5E-14	&9.9E-05 \\
			\#M5	&-			&-			&-			&7.3E-06	&4.5E-14	&9.9E-05 \\
			\#M6	&-			&-			&-			&5.5E-09 	&6.5E-05	&\textcolor{red}{0.17} \\
			\#M7	&-			&-			&-			&2.7E-11 	&1.7E-12	&0.0030 \\
			\hline										 
		\end{tabular}
	\end{center}
	\label{tab:p_values}
\end{table}

\subsection{Alpha Matte Labelling}
The alpha mattes are labelled in a semi-automatic way, i.e., rough mattes are generated by image matting methods and refined manually by image editors to fit the anatomical structure, which is efficient and widely used in natural image matting~\cite{2016Deep}. 

Here, we use Information-Flow~\cite{aksoy2017designing}, a laplacian-based matting method, to create the rough alpha mattes as its comparatively better performance in our scenarios, while other methods are also alternatives. 
A trimap is required in image matting methods, which denotes the foreground, background, and unknown region as prior information. We generate it by the manual labelled masks. 
Specifically, pixels are categorized as foreground or background only if tagged the same label in all the binary masks. The remained inconsistent pixels are tagged as the unknown region. Then, we perform morphological dilation to expand the uncertain regions to mitigate the misclassification of foreground and background introduced by boundary blurring in manual labelling.

Three qualified clinicians\footnote{The clinicians involved in the evaluation were all from tertiary hospitals in China, including pediatricians and oncologists.} were invited to review the various manual labels, including the proposed alpha matte and the traditional binary masks. Each of them chose labels to describe the anatomical structures best, shown in Fig.~\ref{fig:vote}. 
We also quantitatively analyzed whether alpha matte was more popular by t-tests, and the results are shown in the Table~\ref{tab:p_values}.
It is demonstrated that the alpha mattes are significantly more favorable than the binary masks.

\section{Methodology}\label{sec:methods}

\begin{figure*}[t]
	\begin{center}
		\includegraphics[width=\textwidth]{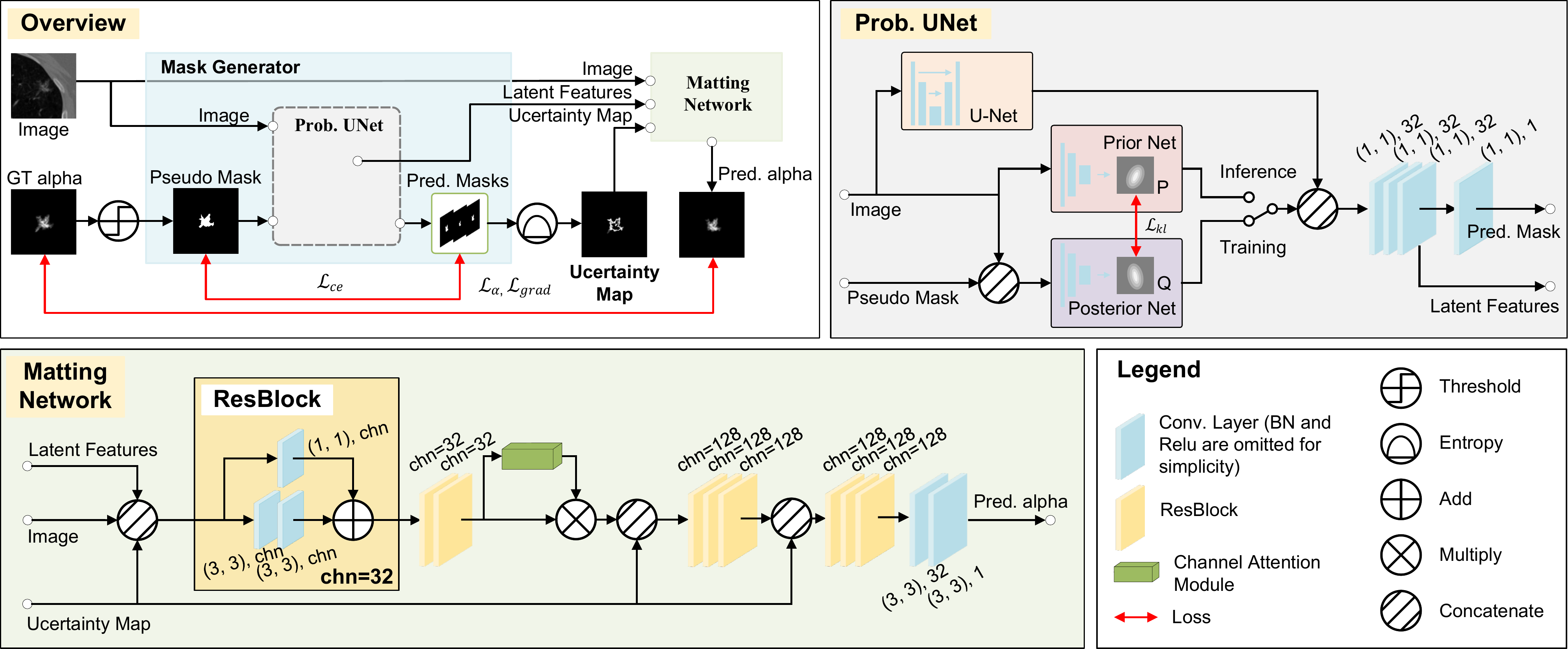}
	\end{center}
	\caption{The schematic diagram of Medical Matting. The Mask Generator outputs various segmentation predictions under the target distribution by Prob. UNet. These intermediate results are merged to create an Uncertainty Map, which assists the following Matting Network to predict the alpha matte. The generation of binary masks simulates the process of doctor's labeling. Using the corresponding result differences, we obtain the uncertain region for future use in the matting network to reduce the training difficulty. We show the details of layers in the form \textit{(kernel size), channel} for understanding the network and re-implementation. 
	}
	\label{fig:structure}
\end{figure*}

Labeling with continuous values can accurately depict the anatomical structure and provide an intuitive way to quantify uncertainty as shown in Section~\ref{sec:introduction} and~\ref{sec:datasets}.
In this section, we introduce a multi-task network which generates binary masks and alpha matte together. 

The network consists of two closely integrated parts, the \textbf{mask generator} used for binary segmentation and the \textbf{matting network} used for predicting alpha mattes.
The binary masks predicted by the generative model can be regarded as the simulation of clinicians' labeling. 
We use the \textbf{uncertainty map}, metricizing the difference by entropy among the multiple mask predictions, to obtain the region concerned, and feed it to the matting network as an auxiliary for predicting alpha matte. 
Fig.~\ref{fig:structure} provides a schematic view of the framework.

\subsection{Mask generator}
Mask generator is a generative model to produce a bunch of binary masks, and the intermediate score maps are used to build an uncertainty map as auxiliary information to assist the training of the matting network. 
For simplicity and without loss of generality, Prob. UNet\footnote{We refer to the Pytorch implementation, which can be achieved publicly at: \color{blue}{https://github.com/stefanknegt/Probabilistic-Unet-Pytorch}.}~\cite{kohl2018probabilistic}, rather than its more complex descendants, was chosen in our network to create a set of binary masks under the target distribution. 

Since the biological tissue structures are generally continuous, this continuity will generally be reflected in the captured images and, in turn, in the corresponding alpha mattes.
Therefore unlike the original method, which randomly samples a mask from multiple labelled masks in each training iteration, we generate a pseudo binary mask by random thresholding the ground truth alpha matte.
Thus we can generate more abundant masks with structural continuity, which is advantageous over the limited manually labelled masks.
In addition, masks generated by different thresholds correspond to different uncertainty tolerances.
The pseudo mask $\widetilde{m}$ can be generated by Eq.~\ref{eq: mask_generator}. 
\begin{equation}
	\widetilde{m}(x)=
	\begin{cases}
		1       & \quad {\boldsymbol{\alpha}_{gt}(x) \geq \tau}\\
		0  		& \quad {\boldsymbol{\alpha}_{gt}(x) < \tau}
	\end{cases}, 
	\tau\in \left [a, b\right]
	\label{eq: mask_generator}
\end{equation}
where $\boldsymbol{\alpha}_\mathit{gt}$ denotes the ground truth alpha matte, $\tau$ stands for the threshold level, $a$ and $b$ define the range of $\tau$, and are practically set to 0.2 and 0.7 of the maximum of $\boldsymbol{\alpha}_\mathit{gt}$ to get reasonable masks.


\subsection{Uncertainty map}

Since matting requires both a clear distinction between the target and the background and a description of the target region using continuous values, it is a task involving both classification and regression. Consequently,  it is a challenge to perform matting directly by regressing every pixel. Thus, trimap is introduced as a priori constraint that indexes the foreground, background, and unknown regions, vastly reducing the task complexity and is widely used as a consensus in both laplacian-based and deep learning-based methods~\cite{aksoy2017designing, cai2019disentangled, chen2013knn, chuang2001bayesian, forte2020fbamatting}.
Given this, we propose the uncertainty map as a similar auxiliary structure in our network.

Unlike natural images, sometimes there is no definite foreground, or it is difficult to distinguish the foreground area (i.e., the lesions) from its surrounding structures in medical images, e.g., a CT slice through a ground glass-like area around a pulmonary nodule.
Therefore, the trimap mechanism cannot be applied directly.

Inspired by the Monte Carlo Dropout approximation~\cite{kendall2017uncertainties}, we have created a score map named \textit{uncertainty map}, generated by the predictions of Prob. UNet. 
The uncertainty map $\mathcal{U}_\mathit{map}$ indicates the challenging areas in continuous values to be identified and plays a similar role to the trimap, which is defined as the entropy:
\begin{equation}
	\mathcal{U}_\mathit{map}(x_i)=-\sum_{c=1}^C\overline{p}^c(x_i)log\overline{p}^c(x_i) \label{eq: Uncertainty map}
\end{equation}
where $C$ is the number of classes, and $\overline{p}^c(x_i)$ is the probability of the pixel $x_i$ in class $c$ of the average score map of the Prob. UNet predictions. 
Suppose we generate $N$ score maps \{$\hat{p}_1^c, \hat{p}_2^c, \cdots, \hat{p}_N^c$\} per image class, then $\overline{p}^c(x_i)=\frac{1}{N}\sum_{n=1}^N \hat{p}_n^c(x_i)$. 

In this way, we obtain an initial description of uncertainty in the data, which is more expressive than trimap with continuous values. 
At the same time, it is fully data-driven, and it can better adapt to upstream and downstream tasks in network training.
Fig.~\ref{fig:entropy}-(c) shows examples of the generated uncertainty maps, from which we can see that areas that are easy to be confused, such as the edges of the lesions, are highlighted. 


\subsection{Matting network}
The matting network generates the final alpha matte prediction with the help of the uncertainty map. 
It consists of three propagation units, each of which consists of three residual blocks~\cite{he2016deep}. 
A channel attention module~\cite{woo2018cbam} is inserted between the first two units to help the network pay attention to effective features. 
The output block consists of two convolution layers at the end of the pipeline. 
The input image, latent features from the Prob. UNet, and the uncertainty map are concatenated as the matting network's input. 
The uncertainty map is also applied to the last two propagation units providing constraint information~\cite{cai2019disentangled}.
The detailed network structure can refer to Fig.~\ref{fig:structure} or the code.


\subsection{Multi-tasking loss weighting strategy}
The binary mask prediction and alpha mette prediction form a multitasking network that provides facilitated end-to-end training. 
Moreover, the information shared between related tasks can achieve better performance~\cite{girshick2015fast}. 
In this section, we first introduce the losses used in each task, and then introduce the weight balancing strategy of the two tasks.

\subsubsection{Losses}
The total training loss consists of two parts, $\mathcal{L}_\mathit{seg}$ for binary mask prediction and $\mathcal{L}_\mathit{matt}$ for alpha matte.

For binary mask prediction, cross-entropy loss $\mathcal{L}_\mathit{ce}$ and Kullback-Leibler loss $\mathcal{L}_\mathit{kl}$ are applied.
The former is for matching the predicted mask and the pseudo ground truth mask, and the latter is for minimizing divergence between the prior distribution $P$ and the posterior distribution $Q$~\cite{kohl2018probabilistic}. 
$\mathcal{L}_\mathit{ce}$ and $\mathcal{L}_\mathit{kl}$ are defined as:
\begin{equation}
	\mathcal{L}_{ce}\left (\hat{p}^c, \widetilde{m}^c\right )=
	- \frac{1}{N}\sum_{i=1}^N\sum_{c=1}^C
	 \widetilde{m}^c(x_i) \log \hat{p}^c(x_i)
	\label{eq: ce_loss}
\end{equation}
where $\hat{p}^c$ and $\widetilde{m}^c$ are the $c^{th}$ channel of the predicted score map from mask generator network and pseudo ground truth mask, respectively. $x_i$ is the $i^{th}$ pixel in the image. $N$ stands for the number of total pixels in the image.
\begin{equation}
	\mathcal{L}_{kl}(Q\|P)=
	\mathbb{E}_{z\sim Q}\left[
		\log Q(z|\widetilde{\mathcal{M}}, \mathcal{X}) - \log P(z| \mathcal{X}))
	\right]
	\label{eq: kl_loss}
\end{equation}
where $z$ is a sample from the distribution, $\mathcal{X}$ and $\widetilde{\mathcal{M}}$ are for the input images and pseudo ground truth masks, respectively.
Then we have $\mathcal{L}_\mathit{seg} = \mu\mathcal{L}_\mathit{ce}+\upsilon\mathcal{L}_\mathit{kl}$, in which  $\mu$, $\upsilon$ are the weighting parameters.

For matting, the absolute difference and the gradient difference between the predicted alpha matte and the ground truth alpha matte are both considered by $\mathcal{L}_\mathit{\alpha}$ and $\mathcal{L}_\mathit{grad}$, respectively.
The gradient indicates the correlation between a pixel and its surrounding pixels, which is important to the medical structure's continuity.
Moreover, a mask generated by the uncertainty map is applied to concentrate the gradient loss in the uncertain regions.
The losses are defined as:
\begin{equation}
	\mathcal{L}_{\boldsymbol{\alpha}}\left (\hat{\boldsymbol{\alpha}}, \boldsymbol{\alpha}_\mathit{gt}\right )=
	\frac{1}{N}\sum_{i=1}^N
	\left \|\hat{\boldsymbol{\alpha}}\left(x_i\right)-\boldsymbol{\alpha}_\mathit{gt}\left(x_i\right)\right \|_1  \label{eq:alpha_loss}
\end{equation}
\begin{equation}
	\mathcal{L}_{grad}\left ( \hat{\boldsymbol{\alpha}}, \mathcal{U}_\mathit{map}, \boldsymbol{\alpha}_\mathit{gt}\right )=
	\frac{1}{\sum \mathbbm{1}_{\mathcal{R}}}\sum_{x_i \in \mathcal{\mathcal{R}}}
	\left \|\nabla_{\hat{\boldsymbol{\alpha}}}(x_i)-\nabla_{\boldsymbol{\alpha}_\mathit{gt}}(x_i)\right \|_1  
	\label{eq:alpha_grad_loss}
\end{equation}
where $\hat{\boldsymbol{\alpha}}$, $\boldsymbol{\alpha}_\mathit{gt}$ denote the predicted and ground truth alpha matte, respectively. A sub-region $\mathcal{R}$ of $\mathcal{U}_\mathit{map}$ selected by thresholding $\tau$ is used as a mask for gradient loss, defined as $\mathcal{R}=\mathcal{U}_\mathit{map}>\tau$, making the loss more focused in the uncertain regions. The overall matting loss is then given by $\mathcal{L}_\mathit{matt}=\zeta\mathcal{L}_\mathit{\boldsymbol{\alpha}}+\xi\mathcal{L}_\mathit{grad}$, where $\zeta$ and $\xi$ are coefficients.

\subsubsection{Loss weighting strategy}
The performance of multi-task learning is critically influenced by the relative weights of subtasks~\cite{gong2019comparison}. Therefore, an uncertainty-based weight assignment scheme~\cite{kendall2018multi} was proposed, which is also the one we adopted in the conference paper. 
In this paper, according to our specific task, an oscillation attenuation weighting strategy is proposed to improve performance further. 

\textbf{Uncertainty weighting strategy}~\cite{kendall2018multi} (UWS) treats the outputs of each task following a Gaussian distribution with observation noise (for example, $\sigma_1$ and $\sigma_2$ for two subtasks scenario).
The losses of subtasks can be balanced by homoscedastic task uncertainty.
With mathematical derivation of maximising the log-likelihood of the model and minimizing the objective function, the overall loss with UWS $\mathcal{L}_\mathit{UWS}$ can be written as:
\begin{equation}
	\mathcal{L}_\mathit{UWS}=\frac{1}{\sigma_1^2} \mathcal{L}_\mathit{seg}
	+ \frac{1}{2\sigma_2^2}\mathcal{L}_\mathit{matt}
	+ \log\sigma_1\sigma_2 
	\label{eq:multi-loss U}
\end{equation}
where $\sigma_1$ and $\sigma_2$ are trainable parameters.

\begin{figure}
	\centering
	\includegraphics[width=\columnwidth]{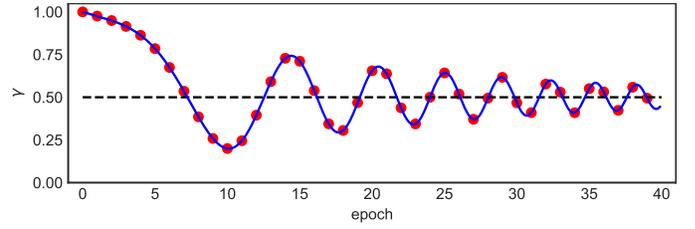}
	\caption{Illustration of the weights change in oscillation attenuation weighting strategy, in which $a=0.05$, $b=0.03$, $t=0.50$.}
	\label{fig:loss_strategy}
\end{figure}

\textbf{Oscillation attenuation weighting strategy} 
(OAWS) controls the weights of loss terms with a periodic function $\gamma$.
In our task scenario, the matting network is dependent on the results of the segmentation network, which is different from other multi-task learning networks.  Inspired by stage-wise training~\cite{barshan2015stage}, we adjust subtask weights cyclically so that the network training focuses on the preorder task first and then dynamically and steadily switches the training focus. Moreover, we can see each oscillation cycle as a coarse-to-fine optimization stage. The weight assigned to $\mathcal{L}_\mathit{seg}$ is given by
\begin{equation}
	\begin{aligned}
		\gamma = \frac{1}{2}g(n)\cos(nf(n)) + t
	\end{aligned}
	\label{eq:weighting_strategy}
\end{equation}
where $n$ stands for the index of epoch, $g(n)$ is a monotonically decreasing function used to control the decay rate of the weights, $f(n)$ is used to control the oscillation period, and $t$ is the final convergence target of the weight. For simplicity, we take $g(n)=e^{-an}$, $f(n)=bn$, where $a$ and $b$ are coefficients. Then the total loss with OAWS $\mathcal{L}_\mathit{OAWS}$ is defined as:
\begin{equation}
	\mathcal{L}_\mathit{OAWS}=\gamma \mathcal{L}_\mathit{seg}
	+ (1-\gamma)\mathcal{L}_\mathit{matt}
	\label{eq:multi-loss OA}
\end{equation}
An example of how the weights change as the epoch increases is shown in Fig.~\ref{fig:loss_strategy}.
\section{Experiments}\label{sec:experiments}

In Section~\ref{sec:datasets}, we have described and illustrated the technical details and advantage of alpha matte for characterizing lesions. 
In this part, we will focus on demonstrating the effectiveness of the network through various experiments. 
Since our work is the first to use deep image matting to handle uncertainty, there exist no models for a direct and comprehensive comparison.
Therefore, we have re-implemented  six state-of-the-art matting methods and compare our proposed method with them quantitatively and qualitatively. Moreover, we also elucidate the role of the proposed uncertainty map and multi-task learning through ablation experiments.


\subsection{Implementation details}

\begin{table}[!t]
	\caption{Hyper-parameters for training}
	\begin{center}
		\begin{tabular}{l|ccc}
			\hline
			Hyper-params	&LIDC-IDRI~\cite{armato2004lung}   &ISIC~\cite{codella2018skin}	&Brain-growth~\cite{https://qubiq.grand-challenge.org} \\			
			\hline
			base $l_r$	&$5\times 10^{-4}$	&$1\times 10^{-4}$	&$1\times 10^{-4}$ \\
			epoch	&80	&100	&150 \\ 
			ipt. Size	&$128 \times 128$	&$256 \times 256$	&$128 \times 128$ \\ 
			batch size	&32	&8	&4 \\
			optimizer	&ADAM	&ADAM	&ADAM \\		
			weight decay	&$5\times10^{-5}$	&$5\times10^{-5}$	&$5\times10^{-5}$ \\
			momentum	&0.9	&0.9 &0.9 \\
			$[\mu, \upsilon, \zeta, \xi]$ &$[1, 10, 1, 1]$	&$[1, 10, 1, 1]$	&$[1, 10, 1, 1]$ \\ 
			\hline										 
		\end{tabular}
	\end{center}
	\label{tab:training_details}
\end{table}

\begin{table*}[!t]
	\caption{Quantitative comparisons with state-of-the-art matting algorithms. 
	}
	\begin{threeparttable}
		\centering
		\begin{tabular}{l|llll|llll|llll}
			\hline
			Datasets& \multicolumn{4}{c|}{LIDC-IDRI}	& \multicolumn{4}{c|}{ISIC}	& \multicolumn{4}{c}{Brain-growth} \\
			\hline
			Model	&	SAD↓   & MSE↓   & Grad.↓ & Conn.↓ &	SAD↓   & MSE↓   & Grad.↓ & Conn.↓ &	SAD↓   & MSE↓   & Grad.↓ & Conn.↓ \\			
			
			Bayesian~\cite{chuang2001bayesian}			& 0.0778 & 0.0819 & 0.1535 & 0.0725	& 7.7535 & 0.1624 & 9.2632 & 7.8800	& 0.8435 & 0.1662 & 1.5921 & 0.8684 \\
			Closed-Form~\cite{levin2007closed}			& 0.3040 & 0.4736 & 0.7584 & 0.3189 & 21.7274 & 0.9062 & 2.7009 & 22.3071 & 1.5419 & 0.4410 & 2.6960 & 1.6259 \\
			KNN~\cite{chen2013knn}						& 0.0737 & 0.0451 & 0.1381 & 0.0732 & 7.6282 & 0.1861 & 4.1263 & 7.7117 & 0.6534 & 0.1073 & 1.1548 & 0.6945 \\
			Information-Flow~\cite{aksoy2017designing}	& 0.0663 & 0.0351 & 0.1001 & 0.0654 & 5.3062 & 0.1061 & 2.8643 & 5.3037 & 0.6819 & 0.1056 & 1.5007 & 0.7210 \\
			Learning Based~\cite{zheng2009learning}		& 0.0554 & 0.0286 & 0.0826 & 0.0509 & 8.4567 & 0.2113 & 4.8210 & 8.6307	& 0.6061 & 0.0898 & 1.0559 & 0.6443 \\
			FBA~\cite{forte2020fbamatting}				& 0.0598 & 0.0395 & 0.1143 & 0.0558 & 8.8235 & 0.2590 & 4.9446 & 8.9998 & 0.7711 & 0.1390 & 1.2350 & 0.8229 \\
			Medical Matting (Ours)						& \bf{0.0447} & \bf{0.0215} & \bf{0.0607} & \bf{0.0378} & \bf{1.0330} & \bf{0.0093} & \bf{0.1729} & \bf{0.4989} & \bf{0.4023} & \bf{0.0451} & \bf{0.5572} & \bf{0.4255} \\
			\hline										 
		\end{tabular}
	\begin{tablenotes}
		\item SAD, Grad., and Conn. are scaled by $1\times 10^{-3}$. 
		We only evaluated the uncertain regions because some algorithms optimized only such regions.
    \end{tablenotes}
	\end{threeparttable}
	\label{tab:result-matting}
\end{table*}

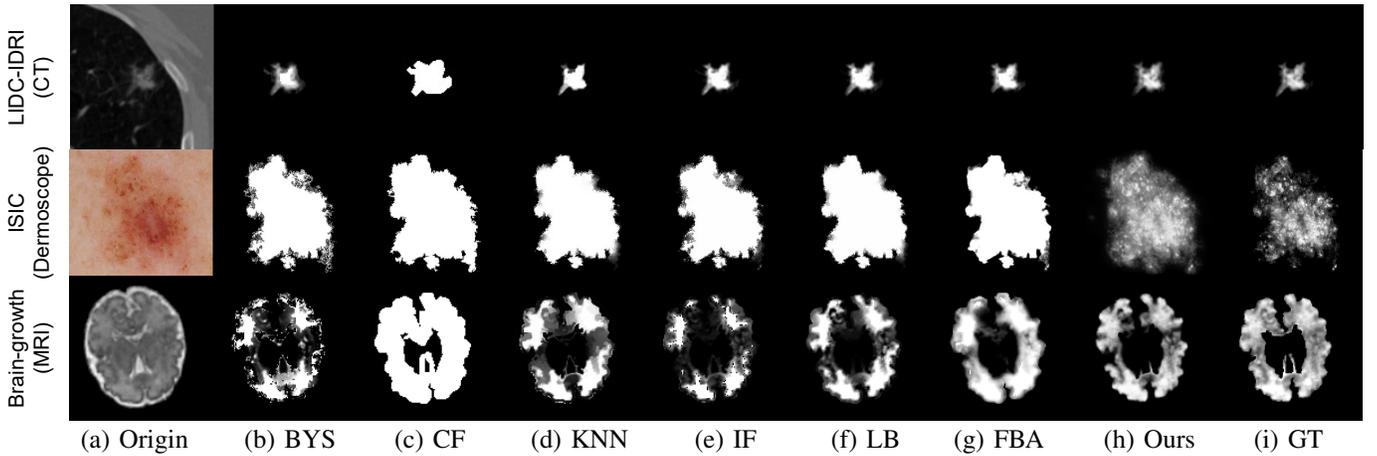
\begin{figure*}[!t]
	\begin{center}
		\begin{overpic}[width=\textwidth]{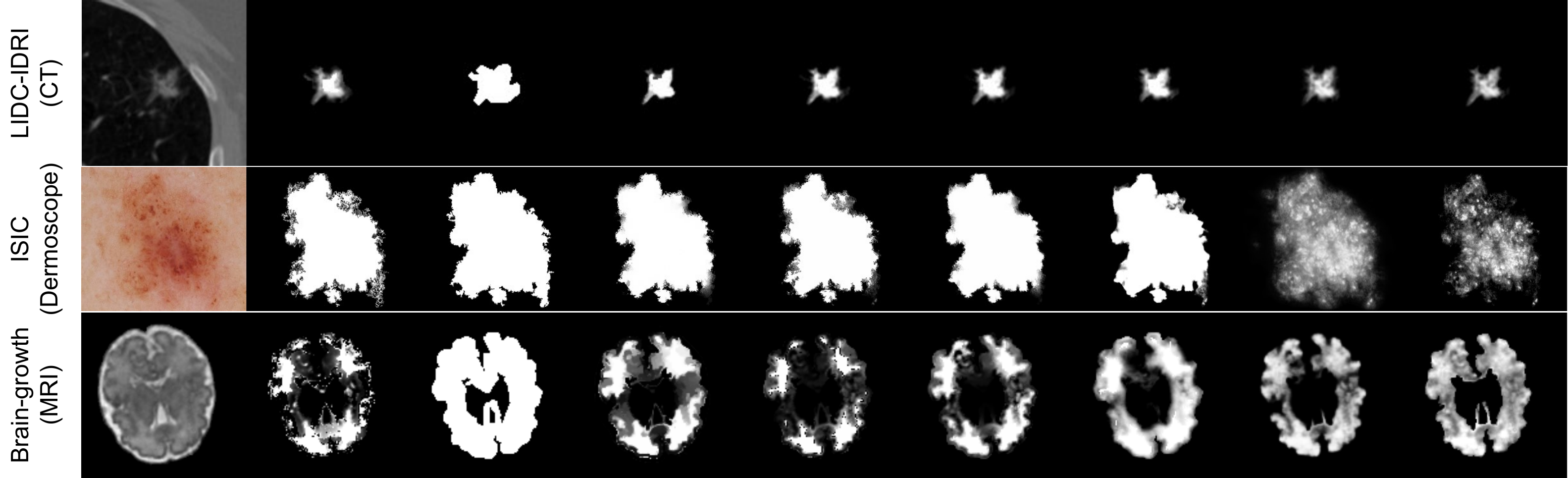}
			\put(6,-2){(a) Origin}
			\put(18,-2){(b) BYS}
			\put(29,-2){(c) CF}
			\put(39,-2){(d) KNN}
			\put(51,-2){(e) IF}
			\put(61,-2){(f) LB}
			\put(70,-2){(g) FBA}
			\put(81,-2){(h) Ours}
			\put(92,-2){(i) GT}
		\end{overpic}
	\end{center}
	\caption{Qualitative comparisons on samples from LIDC-IDRI, ISIC, and Brain-growth datasets. 
	Origin: original images, BYS: Bayesian~\cite{chuang2001bayesian}, CF: Closed-Form~\cite{levin2007closed}, KNN~\cite{chen2013knn}, IF: Information-Flow~\cite{aksoy2017designing}, LB: Learning Based~\cite{zheng2009learning}, FBA~\cite{forte2020fbamatting}.
	On the medical datasets tested, our method better describes the details of the lesions.
	We show more samples in the supplementary files.
    Better view in zoom and color.
	}
	\label{fig:results}
\end{figure*}

The proposed method is evaluated on the three datasets introduced in Section~\ref{sec:datasets} and four-fold cross-validation is used to reduce the interference caused by the random errors.
Data augmentation are used in data pre-processing, including flip, rotation, and elastic transformation~\cite{simard2003best}.
The threshold for the mask in $\mathcal{L}_\mathit{grad}$ (Eq.~\ref{eq:alpha_grad_loss}) is set to $0.1$. Eight masks are generated for uncertainty map generation. $a, b, t$ in Eq.~\ref{eq:weighting_strategy} are set to $0.05, 0.03, 0.50$, respectively.
All the models are trained from scratch.
The cosine annealing schedule~\cite{bochkovskiy2020yolov4,loshchilov10sgdr} was used after a 1-epoch long steady increasing warm-up from $0$ to base $l_r$. 
The key hyper-parameters are listed in the Table~\ref{tab:training_details}.

Four commonly used evaluation metrics\cite{rhemann2009perceptually} in matting scenes, i.e., absolute differences (SAD), mean squared error (MSE), gradient (Grad.), and connectivity (Conn.) are used to compare the results of alpha matting. 
SAD and MSE are metrics that directly assess the difference between the predicted and the ground truth alpha matte. In contrast, Grad. and Conn. focus on the continuity of the predictions. More specifically, Grad. is an assessment of the difference between two predictions on the gradient, and Conn. focuses on the connectivity between pixels and the largest foreground region.


\subsection{Main results}

Alpha matte can better characterize lesions and be equivalently mapped to the proposed uncertainty. Here we explore the performance of our model in predicting alpha matte qualitatively and quantitatively. Experiments are conducted on the three datasets of different modalities to show the generality.

We compared the predicted alpha mattes with six state-of-the-art matting methods, including a bayesian-based method (Bayesian~\cite{chuang2001bayesian}), four laplacian-based methods (Closed-Form~\cite{levin2007closed}, KNN~\cite{chen2013knn}, Information-Flow~\cite{aksoy2017designing}, Learning Based~\cite{zheng2009learning}) and a deeplearning based method (FBA~\cite{forte2020fbamatting}). 
Since these methods require trimap as an input, we provide them with the trimaps we used while generating the ground truth alpha mattes in Section~\ref{sec:datasets}. 
Deep matting model training requires a large number of samples. 
Generally, it uses the dynamic random pasting of the foreground onto a background to generate a large number of samples~\cite{xu2017deep,forte2020fbamatting} as a data augmentation mechanism. However it is challenging for medical scenarios due to the limitation of the sample size of the dataset. And also this simple paste can not guarantee the correctness of the generated images in terms of anatomical structure. 
Therefore, we evaluate the performance of the FBA using a model trained on natural images.

Table~\ref{tab:result-matting} and Fig.~\ref{fig:results} show the quantitative and qualitative comparison of the results, respectively.
Table~\ref{tab:result-matting} shows that our model outperforms all the other methods in all three datasets, which proves that our method is more applicable to the medical scenarios. 
Our method can better express the edge of the fuzzy transition zone and subtle structural features in the matting results. 
The differences between the foreground and background of medical images are sometimes less prominent than that in natural scenes, and even sometimes the foreground area is hard to give a precise range, and the non-foreground component in the foreground leads to the failure of the trimap mechanism. On the other hand, our proposed uncertainty map is dynamically generated without clearly distinguishing between foreground, background, and uncertainty region, which is more adaptable. In addition, compared with natural images, the high noise level in medical images is one of the reasons affecting the performance of traditional methods. 

Moreover, from the point of view of the model input, our method constructs the uncertainty map by generative model instead of trimap with explicit input, making the method more concise and more automated.


\subsection{Ablation study}

\begin{table*}[!t]
	\caption{Ablation study of multitasking learning and uncertainty map. 
	}
	\begin{threeparttable}
	\centering
		\begin{tabular}{c|ccc|cc|ccc}
			\hline
			Dataset	&Matting Network & Uncertainty map & Loss Strategy & $D_{GED}$↓ & Dice↑ & SAD↓ & MSE↓ & Grad.↓ \\ 
			\hline
			\multirow{7}{*}{LIDC-IDRI}
			&\XSolidBrush (Prob. UNet)	&-	&-	&.2159±.0116	&.9367±.0108	&-	&-	&-	\\
			&\Checkmark	&	 & None 	&.2239±.0155	&.9360±.0105	&.0992±.0100	&.0011±.0002	&.0906±.0162 \\
			&\Checkmark	&\Checkmark	 & None	&\textcolor{blue}{.2183±.0134}	&\textcolor{blue}{.9382±.0102}	&\textcolor{blue}{.0982±.0078}	&.0011±.0002	&\textcolor{blue}{.0887±.0139} \\
			&\Checkmark	&	 & UWS 	&.2257±.0169	&.9338±.0121	&.0981±.0085	&\underline{.0010±.0001}	&.0906±.0125 \\
			&\Checkmark	&\Checkmark	 & UWS 	&\textcolor{blue}{.2230±.0121}	&\textcolor{blue}{.9375±.0119}	&\textcolor{blue}{\underline{.0976±.0068}}	&\underline{.0010±.0001}	&\textcolor{blue}{\underline{.0881±.0107}} \\
			&\Checkmark	&	 & OAWS 	&.2146±.0126	&.9369±.0111	&.0983±.0088	&.0011±.0002	&.0887±.0123 \\
			&\Checkmark	&\Checkmark	 & OAWS	&\textcolor{blue}{\underline{.2146±.0108}}	&\textcolor{blue}{\underline{.9390±.0101}}	&\textcolor{red}{.1007±.0098}	&\textcolor{red}{.0012±.0003}	&\textcolor{red}{.0942±.0185} \\ 	
			\hline
			\multirow{7}{*}{ISIC} 
			&\XSolidBrush (Prob. UNet)	&-	&-	&.3559±.0696	&.8494±.0480	&-	&-	&-	\\
			&\Checkmark	&	 & None 	&.3043±.0752	&.8711±.0485	&2.2825±.7631	&.0096±.0062	&.4631±.1700 \\
			&\Checkmark	&\Checkmark	 & None	&\textcolor{red}{.3245±.0619}	&\textcolor{red}{.8611±.0468}	&\textcolor{blue}{2.2652±.6777}	&\textcolor{red}{.0099±.0050}	&\textcolor{red}{.4704±.1447} \\
			&\Checkmark	&	 & UWS 	&\underline{.2986±.0774}	&\underline{.8743±.0512}	&2.3178±.8065	&.0101±.0063	&.4597±.1840 \\
			&\Checkmark	&\Checkmark	 & UWS 	&\textcolor{red}{.3042±.0772}	&\textcolor{red}{.8636±.0584}	&\textcolor{blue}{2.0557±.6483}	&\textcolor{blue}{.0078±.0047}	&\textcolor{blue}{.4223±.1388} \\
			&\Checkmark	&	 & OAWS 	&.3131±.0862	&.8633±.0572	&2.1345±.7301	&.0083±.0054	&.4207±.1621 \\
			&\Checkmark	&\Checkmark	 & OAWS	&\textcolor{red}{.3228±.0723}	&\textcolor{red}{.8570±.0517}	&\textcolor{blue}{\underline{1.8016±.6073}}	&\textcolor{blue}{\underline{.0060±.0040}}	&\textcolor{blue}{\underline{.4037±.1554}} \\ 	
			\hline
			\multirow{7}{*}{Brain-growth} 
			&\XSolidBrush (Prob. UNet)	&-	&-	&.3388±.0478	&.8916±.0219	&-	&-	&-	\\
			&\Checkmark	&	 & None 	&.3384±.0356	&.8971±.0127	&.6412±.0634	&.0127±.0029	&.9393±.2119 \\
			&\Checkmark	&\Checkmark	 & None	&\textcolor{red}{.3448±.0529}	&\textcolor{red}{.8965±.0187}	&\textcolor{blue}{.6014±.0575}	&\textcolor{blue}{.0114±.0023}	&\textcolor{blue}{.8796±.1893} \\
			&\Checkmark	&	 & UWS 	&.3154±.0307	&.9024±.0123	&.5830±.0492	&.0104±.0017	&.8068±.1333 \\
			&\Checkmark	&\Checkmark	 & UWS 	&\textcolor{blue}{.3145±.0437}	&\textcolor{blue}{.9032±.0166}	&\textcolor{red}{.5867±.0558}	&\textcolor{blue}{.0102±.0017}	&\textcolor{blue}{.7879±.1298} \\
			&\Checkmark	&	 & OAWS 	&.3052±.0308	&.9026±.0130	&.6047±.0539	&.0113±.0021	&.8511±.1402 \\
			&\Checkmark	&\Checkmark	 & OAWS	&\textcolor{blue}{\underline{.2994±.0446}}	&\textcolor{blue}{\underline{.9046±.0160}}	&\textcolor{blue}{\underline{.5685±.0398}}	&\textcolor{blue}{\underline{.0101±.0017}} 	&\textcolor{blue}{\underline{.7749±.1384}} \\ \hline						 
		\end{tabular}
		\begin{tablenotes}
			\item SAD, Grad. are also scaled by $1\times 10^{-3}$. The data is expressed in the format of mean±std. If the performances have improved after using the uncertainty map, they are marked in blue and vice versa in red. The best scores are marked with underlines. 
		\end{tablenotes}
	\end{threeparttable}
	\label{tab:ablation}
\end{table*}

In this ablation study, we analyze the gain of multitask-learning, uncertainty map, and loss strategy on the model performance in depth. Specifically, we compare the performance difference in distribution prediction between the multi-task network and the single-task model (Prob. UNet). Also, we compare the differences of three multi-task learning weighting strategies, including None (adding the $\mathcal{L}_\mathit{seg}$ and $\mathcal{L}_\mathit{matt}$ directly), UWS, and OAWS. Moreover, we evaluate the role of the uncertainty map under each strategy. 

Gerneralized energy distance $D_\mathit{GED}$~\cite{kohl2018probabilistic} and a adapted dice $\mathit{Dice}$ are used to evaluate the similarity between the distribution of the predicted and target masks, formulated as:
\begin{equation}
	D_\mathit{GED}(\hat{\mathcal{M}}, \mathcal{M}_\mathit{gt})
	\!=2\mathop{\mathbb{E}}[ d(\hat{m}, m)] \!-\!\mathop{\mathbb{E}}[d(\hat{m}, \hat{m}^{\prime})]
	\!-\!\mathop{\mathbb{E}}[d(m, m^{\prime})]
	\label{eq:U_dice}
\end{equation}
\begin{equation}
	\mathit{Dice}(\hat{\mathcal{M}}, \mathcal{M}_\mathit{gt})=
	\mathop{\mathbb{E}}_{\hat{m}\in \hat{\mathcal{M}}}
	\mathop{max}_{m_\mathit{gt}\in \mathcal{M}_\mathit{gt}}
	\left\lbrace\mathit{dice}(\hat{m}, m_\mathit{gt})\right\rbrace
	\label{eq:max_dice}
\end{equation}
$\hat{\mathcal{M}}$ and $\mathcal{M}_\mathit{gt}$ denote the predicted and target masks, respectively. $\hat{m}, \hat{m}^{\prime}$ are independent samples of the predicted masks, and $m, m^{\prime}$ are independent samples of the target masks. $d(a,b) \!=\! 1\!-\!IoU(a,b)$, where $IoU$ stands for intersection over union. $dice$ refers to the conventional dice coefficient.
$D_\mathit{GED}$ focuses on the similarity evaluation of the distribution as a whole, while $\mathit{Dice}$ focuses more on the similarity evaluation between samples.
Since medical images differ from natural images, there is often more than one foreground target present, and thus Conn. is less applicable when evaluating the whole image. Therefore, we use SAD, MSE, and Grad. here to evaluate matting performance.

For the experiments, we use four-fold cross-validation. 
To ensure that the target masks share the same distribution, we generate 8 binary masks by using Eq.~\ref{eq: mask_generator} on alpha matte with equidistant thresholds, from which a specific number of masks is obtained by evenly sampling.
For UWS, the $\sigma_1$ and $\sigma_2$ are initialized to $4$. 
Coefficients $a$, $b$, and $t$ for OAWS are set to $0.05, 0.03, 0.50$, respectively.
The results are shown in Table~\ref{tab:ablation}.

Overall, the multi-task network with the matting network improves the fitting of the target distribution than the single-task network (Prob. UNet). 
Comparing the weighting strategies, those using the UWS and the OAWS are better than those without them in terms of masking prediction and matting measurement, and OAWS has improved even more. 
Compared with the network without the uncertainty map, the overall performance of the network with the uncertainty map is further improved, and the improvement of SAD and Grad in the matting metrics is more prominent, which reveals the guiding role of the uncertainty map in the blurred areas of the image.
From the perspective of the dataset, these metrics have the most significant magnitude of improvement on Brain-growth, followed by ISIC, which may be related to the size of the dataset, implying that our proposed strategy has more significant improvement for low data regimes. 
Since small datasets are common in medical images, the adaptability of our method to medical images is also demonstrated.

\section{Discussion} \label{sec:conclusion}

\subsection{Multi-task learning}
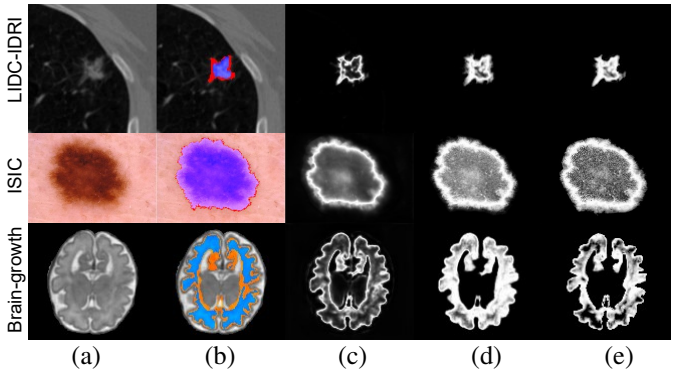
\begin{figure}[!t]
	\begin{center}
		\begin{overpic}[width=\columnwidth]{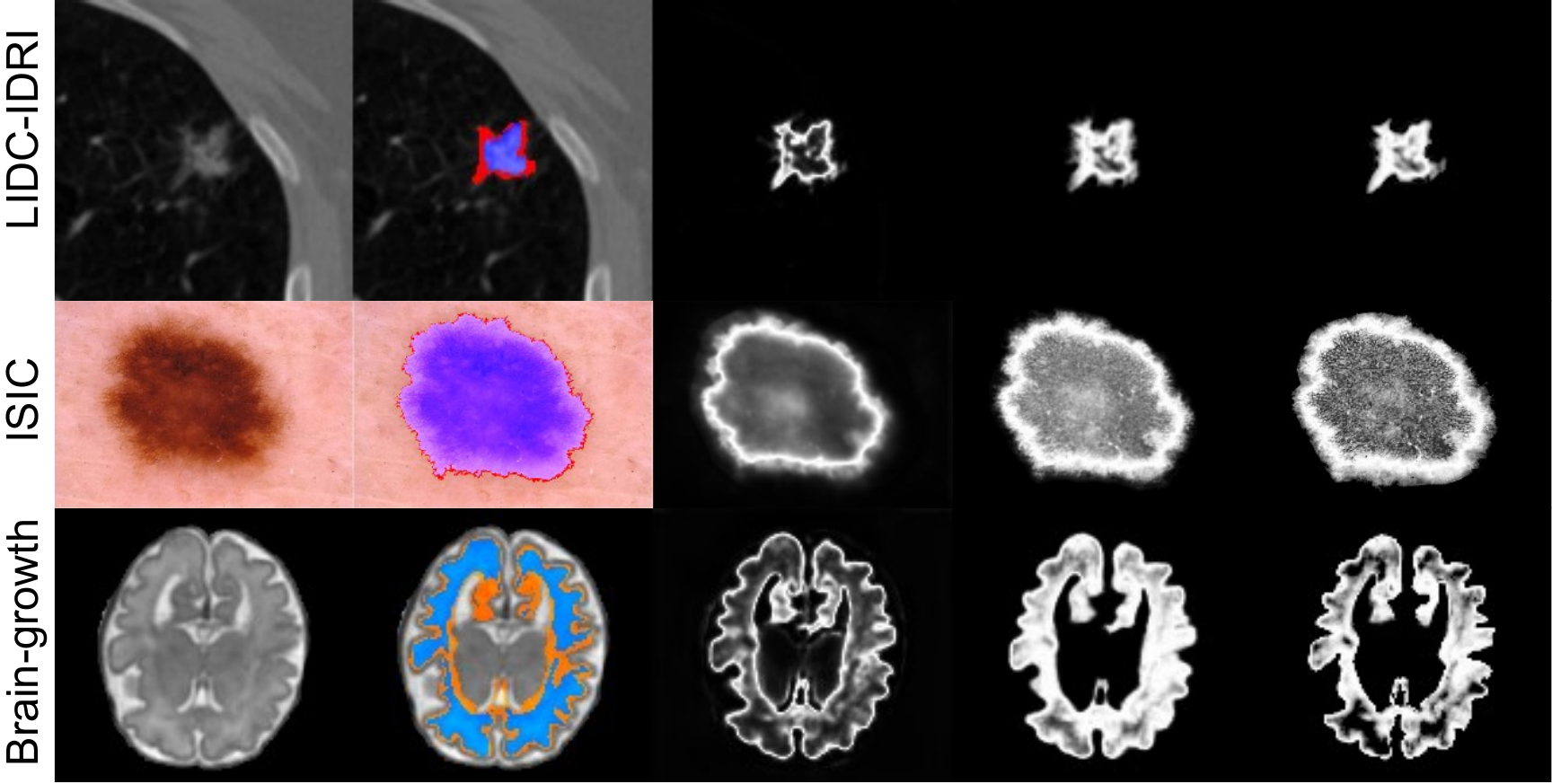}
			\put(10,-3.5){(a)}
			\put(30,-3.5){(b)}
			\put(50,-3.5){(c)}
			\put(70,-3.5){(d)}
			\put(90,-3.5){(e)}
		\end{overpic}
	\end{center}
	\caption{
	Illustration of the coarse-to-fine relationship between the subtasks. 
	The uncertain region obtained by binary masks (The blue and red mask in (b) representing the consistent and inconsistent foreground regions in manual labels, respectively; Uncertainty map (c)) and the entropy of predicted (d) and ground truth (e) alpha mattes are visualized. The results show that the former mainly focuses on the edges of the target in original images (a), while the latter can better represent the internal structure due to the refinement of the matting network.
	}
	\label{fig:entropy}
\end{figure}

By analyzing the correlation between subtasks, we explain the rationals for the performance improvement of multitask network. 
Fig.~\ref{fig:entropy} shows the inconsistent areas of the manual labels, the uncertainty map, and the entropy of the predicted alpha matte (i.e., the proposed uncertainty).
We find that the regions highlighted by the uncertainty map worthy of the matting network's attention are significantly similar to the inconsistent regions between manual annotations, indicating the ability of the uncertainty map to reflect the relatively vague and uncertain regions in images. However, the highlighted parts are focused on the edges of the lesions, while the internal structure of the lesions has not been richly described. In the entropy of the predicted alpha matte, we can see more information of internal structure, which also reflects that the alpha matte can better describe the characteristics of the lesion. Therefore, predicting alpha matte with binary masks can also be viewed as a coarse-to-fine process.

\subsection{Potential use in diagnosis}
To further illustrate that our proposed alpha matte contains more information that can be used for diagnosis than binary masks, we try to feed the classification network by stitching the alpha matte and the input images by channel to observe its improvement in the classification performance.
In addition, we compare different ways of utilizing multiple binary annotations, such as taking the intersection of annotations, the union, the mean, or randomly selecting one of them as the mask in training. To better highlight the gain of mask information for the diagnosis, we choose the lightweight network Resnet18~\cite{he2016deep} as the classifier (two classes, Benign / malignant) and deploy experiments on ISIC dataset. Although the LIDC dataset also has diagnostic labels, the labels are obtained in 3D and thus cannot be directly used in our 2D sliced scenes. We concatenate an all-zero mask to the original image as the baseline to keep the network consistent. 

\begin{figure}
	\centering
	\includegraphics[width=\columnwidth]{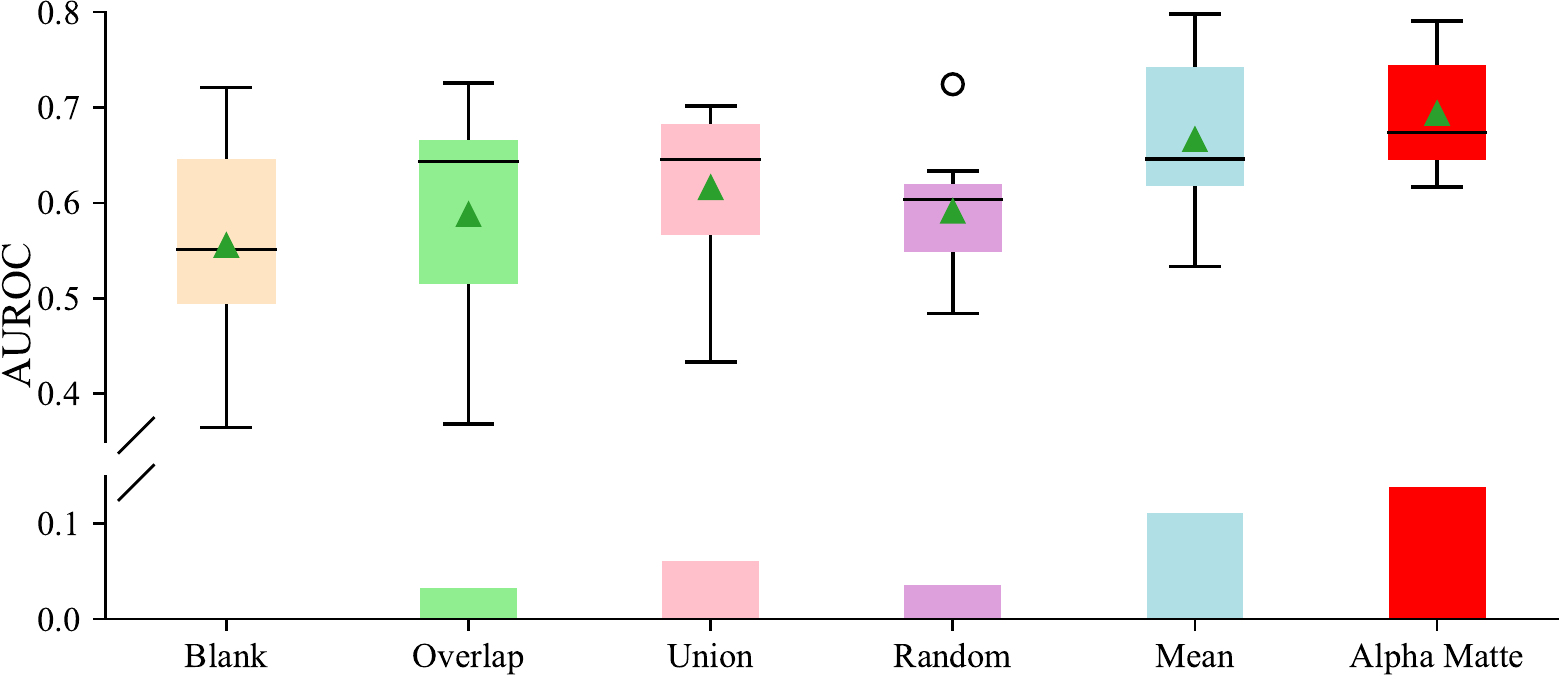}
	\caption{Illustration of the diagnosis performance boost by alpha matte on ISIC dataset. 
	The box plot and bar chart show the AUROCs of different masks combined and their improvements relative to the original input.
	The alpha matte has achieved the greatest improvement.}
	\label{fig:diagnosis}
\end{figure}

Fig.~\ref{fig:diagnosis} shows the diagnosis performance evaluated by Area Under the Receiver Operating Characteristic curve (AUROC) of eight repetitions on ISIC dataset. Alpha matte achieves the best performance improvement while taking the mean value of masks also has a good performance, which indicates that soft labels can indeed provide rich information for diagnosis.

\subsection{Limitation}

Since the alpha matte is manually annotated, even though we try to maintain the consistency of annotation, the dataset we built will still have a certain degree of deviation, which is inevitable due to the subjectivity of labeling and the difference in experience between the annotators. 
However, qualitative analysis, manual evaluation by clinicians, and follow-up experiments show that the performance of the alpha mask is better than that of the binary mask, and as pioneering research, we put this problem on hold for the time being.

\section{Conclusion}

In this work, we creatively calibrate the uncertainty by alpha matte introduced from image matting, which has a greater ability to reveal tiny and ambiguous structures and has a large potential for diagnosis.
A well-designed multi-task network was proposed to predict binary masks and alpha matte simultaneously. 
The uncertainty map, an analogy to trimap, is generated by the intermediate outputs and improves the matting network performance.
The binary masks, uncertainty map, and alpha matte all express the target with uncertainty in different ways, therefore sharing the latent information during training can help the sub-tasks.
The experiments reveal that our model outperforms the other state-of-the-art matting methods on all the four metrics with a considerable margin and demonstrate that alpha matte is a more powerful annotation method than the binary mask.
We labelled three datasets of various modalities with alpha matte, and they were released to the public to promote the study on uncertainty learning and matting in medical scenarios.
We have conducted experiments on data from several modalities, which illustrate the generalizability of our method to a certain extent. However, the current method is applied only to 2D images, while 3D images are also an indispensable part of medical images, providing more abundant structural information. Therefore, the application on 3D images will be the focus of our subsequent research.

\appendices

\section*{Acknowledgment}
The authors would like to thank the clinicians for their hard work in evaluating the alpha matte datasets. They are Yi Luo of Chongqing hospital of traditional Chinese medicine, Huan Luo of Chongqing Renji Hospital of Chinese Academy of Sciences, and Feng Jiang of the First Affiliated Hospital of Wenzhou Medical University.

\bibliographystyle{IEEEtran}
\bibliography{mybibliography}

\end{document}